\documentclass[subscriptcorrection,upint,varvw,barcolor=Goldenrod3,
               mathalfa=cal=euler,balance,hyphenate,french,nolists]{asmejour}

\usepackage{algorithm}
\usepackage{algpseudocode}
\usepackage{float}
\usepackage{caption}
\usepackage[T1]{fontenc}
\usepackage{graphicx}
\usepackage{textcomp}

\JourName{Applied Mechanics}

\begin{document}


\SetAuthorBlock{Kangzheng Liu}{School for Engineering of Matter, Transport Energy,\\
Arizona State University, \\
Tempe, AZ, 85287, USA \\
   email: kangzhen@asu.edu} 

   
\SetAuthorBlock{Leixin Ma\CorrespondingAuthor}{%
School for Engineering of Matter, Transport Energy,\\
Arizona State University, \\
Tempe, AZ, 85287, USA \\
email: leixin.ma@asu.edu
}

\title{MeshODENet: A Graph-Informed Neural Ordinary Differential Equation Neural Network for Simulating Mesh-Based Physical Systems}



\begin{abstract}
The simulation of complex physical systems using a discretized mesh is a cornerstone of applied mechanics, but traditional numerical solvers are often computationally prohibitive for many-query tasks. While Graph Neural Networks (GNNs) have emerged as powerful surrogate models for mesh-based data, their standard autoregressive application for long-term prediction is often plagued by error accumulation and instability. To address this, we introduce MeshODENet, a general framework that synergizes the spatial reasoning of GNNs with the continuous-time modeling of Neural Ordinary Differential Equations. We demonstrate the framework's effectiveness and versatility on a series of challenging structural mechanics problems, including one- and two-dimensional elastic bodies undergoing large, non-linear deformations. The results demonstrate that our approach significantly outperforms baseline models in long-term predictive accuracy and stability, while achieving substantial computational speed-ups over traditional solvers. This work presents a powerful and generalizable approach for developing data-driven surrogates to accelerate the analysis and modeling of complex structural systems.
\end{abstract}

\date{Version \versionno, \today}

\maketitle 


\section{Introduction}
\label{sec:introduction}

The simulation of complex physical systems, often described by partial differential equations on mesh-based discretizations, is fundamental to progress in applied mechanics and various scientific disciplines. While traditional numerical methods like the Finite Element Method and Finite Volume Method offer robust solutions, they frequently incur substantial computational expenses, particularly for large-scale, nonlinear, or multi-scale phenomena requiring long-term integration. This can hinder rapid design cycles and comprehensive parametric investigations, motivating the development of fast and accurate data-driven surrogate models.

However, learning on mesh-based data presents unique challenges for conventional neural networks. Simple vector-based networks, such as Multilayer Perceptrons (MLPs), require fixed-size inputs, making them fundamentally unsuitable for meshes with varying node numbers. Moreover, when mesh data are flattened into a vector for such networks, the essential topological information defining the system's structure is lost. Convolutional Neural Networks (CNNs) are constrained to structured, Euclidean domains and thus fail to generalize to the irregular meshes ubiquitous in computational mechanics.

Graph Neural Networks (GNNs), in contrast, are inherently well-suited for learning on such graph-structured data \citep{BattagliaGNNReview2018}. Founded on the message-passing approach \citep{Gilmer2017NeuralMP}, which aligns with the local nature of physical interactions, GNNs can effectively learn complex physical laws from data. This explicit modeling of interactions allows models like MeshGraphNet to successfully simulate diverse systems by approximating the underlying operators \citep{Pfaff2021, SanchezGonzalez2020, Allen2022PhysicalIO}. The strong relational inductive biases of such structured message-passing have proven advantageous for a range of physics problems, from fluid dynamics \citep{SanchezGonzalez2020, Brandstetter2022MessagePI} to solid mechanics \citep{Fortunato2018LearningSM, Chen2024PredictingDR}, and are generally more suitable than simpler GCNs \citep{Kipf2017SemiSupervisedCW} or spectral GNNs \citep{Bruna2014SpectralNA} for these tasks. Despite their success, these standard GNNs, which typically predict the system's state one discrete step at a time, face a critical challenge: in long-term autoregressive rollouts, prediction errors can accumulate, leading to instability, especially in systems with strong nonlinearities, such as structures with large deformation \citep{LuschMLDynSys2018}.

To address this temporal instability, recent research has explored combining GNNs with Neural Ordinary Differential Equations (NODEs) \citep{Chen2018}, creating hybrid Graph Neural ODE (GNODE) frameworks. NODEs model dynamics by learning the derivative function of a system's state, enabling continuous-time modeling and offering a natural solution to the problem of error accumulation. The synergy between GNNs, which provides the necessary spatial inductive bias for mesh-based systems, and NODEs, which handles the temporal evolution, is powerful \citep{Poli2019GraphCN, Xhonneux2020ContinuousDG}. This has led to a variety of GNODE applications, with notable examples in rigid body dynamics \citep{He2023EGODEAE}, reactive flows \citep{Zhao2023APC}, and multi-agent systems \citep{Huang2023GeneralizingGO}.

However, a closer examination reveals that the application of GNODEs has thus far been concentrated in domains where the underlying graph structure is largely static, such as in traffic forecasting \citep{fang2021spatial,liu2023graph}, social network analysis \citep{zhang2022improving,wen2022social}, and epidemiology \citep{gao2021stan,wan2024epidemiology}. In these areas, the graph topology and the intrinsic properties of nodes and edges remain fixed or evolve only slowly. In stark contrast, the simulation of structural mechanics represents a fundamentally different challenge, naturally falling into the category of \textit{dynamic graphs} \citep{Wu2020ComprehensiveSG}. In these problems, while the graph connectivity typically remains fixed, the geometric and physical attributes of its nodes and edges (e.g., positions, velocities, relative distances) are in constant, tightly-coupled evolution. Despite the intuitive alignment between the continuous-time nature of ODEs and the evolution of such dynamic-graph physical systems, the application of GNODE frameworks in this context remains largely unexplored.

This paper presents a foundational exploration into this promising research direction. We propose \textbf{MeshODENet}, a versatile framework designed specifically to apply the GNODE approach to the long-term simulation of mesh-based physical systems. Our approach synergistically integrates graph-based spatial interaction learning with continuous-time temporal evolution. We posit that for such systems, a physically grounded architecture is paramount. Therefore, we embed the entire GNN, which operates on the current physical state of the mesh in every evaluation, as the derivative function within the NODE formulation. This allows an adaptive ODE solver to continuously evolve the state of the system while perpetually accounting for the up-to-date physical configuration, providing a robust and generalizable solution.

The primary contributions of this work are as follows:
\begin{enumerate}
    \item We introduce a general and versatile GNODE framework specifically for the long-term, stable simulation of complex physical systems, addressing a notable application gap in computational mechanics and providing a robust alternative to problem-specific solutions.
    \item We conduct a foundational study applying the Graph Neural ODE approach to the domain of complex, mesh-based physical simulations. We propose and validate a physically-grounded architecture where the dynamics function is directly coupled with the evolving, dynamic graph state, demonstrating its suitability and stability for long-term prediction.
\end{enumerate}

The effectiveness and generalization capabilities of our framework are demonstrated through a series of case studies in structural mechanics, compared against a standard GNN simulator.

The remainder of this paper is organized as follows. Section~\ref{sec:methodology} elaborates on the proposed \textbf{MeshODENet} framework. Section~\ref{sec:experiments} presents a comprehensive experimental evaluation, Section~\ref{sec:Limitations} discusses limitations and future work, and Section~\ref{sec:Conclusion} provides concluding remarks.

\section{Methodology}
\label{sec:methodology}

This section elaborates on our proposed data-driven framework, \textbf{MeshODENet}, designed to learn and predict the dynamics of a broad class of mesh-based physical systems. To ground our discussion in established principles of computational mechanics, we first use the simulation of a one-dimensional elastic beam as a representative example to illustrate the traditional, physics-based computational approach. This comparison underscores the challenges and computational burdens inherent in conventional methods, which motivate the development of data-driven surrogates such as our proposed model. We will then detail the architecture and operational principles of \textbf{MeshODENet} in the subsequent sections.

\subsection{Data Generation: Deformable Structural Dynamics}
\label{sec:data_generation}

\begin{figure}
    \centering
    \includegraphics[width=0.8\linewidth]{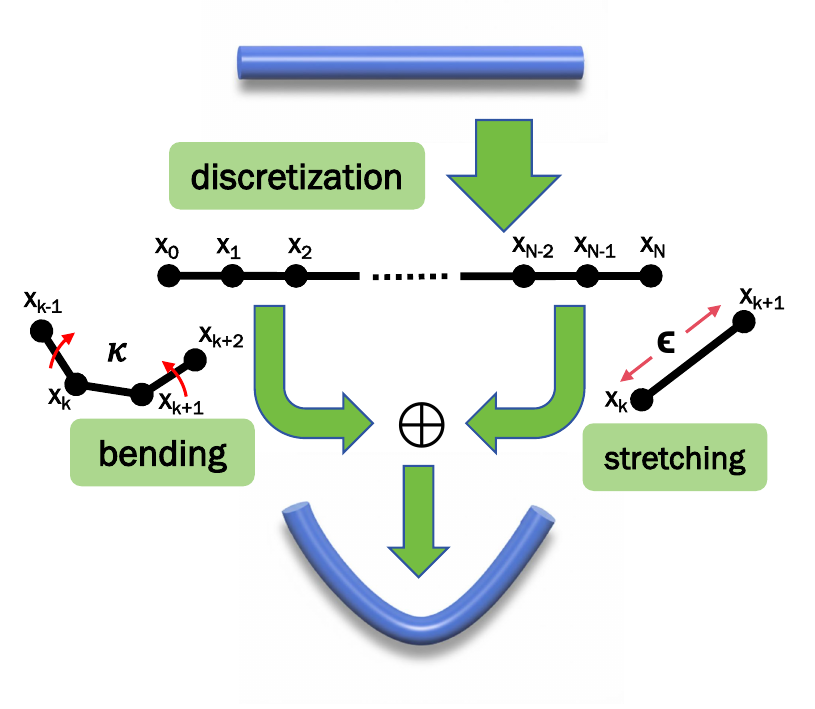}
    \caption{Schematic of the Discrete Elastic Rods (DER) method, illustrating the discretization of a continuous rod and the decomposition into bending ($\kappa$) and stretching ($\epsilon$) components.}
    \label{fig:DER}
\end{figure}

The cornerstone of simulating deformable bodies lies in the spatial discretization of the continuum. The Discrete Elastic Rods (DER) method\citep{DER1,DER2}, founded on Kirchhoff's rod theory and principles of discrete differential geometry, represents a continuous rod as a simplified mechanical system. As illustrated in Fig.~\ref{fig:DER}, the rod’s centerline is discretized into a sequence of $N$ nodes (or vertices), $\mathbf{x}_0, \mathbf{x}_1, \dots, \mathbf{x}_{N-1}$, connected by $N-1$ segments (or edges), with the system's state fully described by the degrees of freedom (DOF) of these nodes. In our motivating example, an elastic beam discretized into 21 nodes undergoes free fall under gravity within a low-Reynolds-number fluid (further discussed in Section~\ref{sec:experiments}).

The dynamics are governed by the elastic potential energy stored in the structure. For each segment, this energy is decomposed into stretching and bending components. The total elastic energy of the rod, $E_{\text{elastic}}$, is the sum over all elements:
\begin{equation}
    E_{\text{elastic}} = \sum_{k=0}^{N-2} E_s^k + \sum_{k=1}^{N-2} E_b^k
\end{equation}

\paragraph{Stretching Energy.} The stretching energy in the $k$-th segment (connecting nodes $k$ and $k+1$) is proportional to the square of the stretching strain, $\epsilon_s$:
\begin{equation}
    E_s^k = \frac{1}{2} (EA) (\epsilon_s^k)^2 l_k^0, \quad \text{where} \quad \epsilon_s^k = \frac{\|\mathbf{x}_{k+1} - \mathbf{x}_k\|}{l_k^0} - 1
\end{equation}
Here, $E$ is the Young's modulus, $A$ is the cross-sectional area, and $l_k^0$ is the rest length of the segment.

\paragraph{Bending Energy.} The bending energy at the $k$-th interior node is proportional to the square of the deviation from the natural curvature, $\kappa_b^0$:
\begin{equation}
    E_b^k = \frac{1}{2} (EI) (\kappa_b^k - \kappa_b^0)^2 l_k
\end{equation}
where $I$ is the second moment of area of the cross-section. The discrete curvature $\kappa_b^k$ measures the turning angle at node $\mathbf{x}_k$ between the adjacent segments $\mathbf{e}_{k-1} = \mathbf{x}_k - \mathbf{x}_{k-1}$ and $\mathbf{e}_k = \mathbf{x}_{k+1} - \mathbf{x}_k$. The vector-valued discrete curvature is computed as:
\begin{equation}
    \boldsymbol{\kappa}_b^k = \frac{2 \mathbf{e}_{k-1} \times \mathbf{e}_k}{\|\mathbf{e}_{k-1}\| \|\mathbf{e}_k\| + \mathbf{e}_{k-1} \cdot \mathbf{e}_k}
\end{equation}
The scalar curvature $\kappa_b^k$ is the magnitude of this vector, and $l_k$ is a local length scale, typically the average length of the two adjacent segments.

The internal elastic forces acting on the nodes are derived as the negative gradient of this total potential energy, $\mathbf{F}_{\text{internal}} = - \partial E_{\text{elastic}} / \partial \mathbf{x}$. This leads to the semi-discretized equation of motion for the system:
\begin{equation}
    \mathbf{M}\ddot{\mathbf{x}} + \mathbf{F}_{\text{internal}}(\mathbf{x}) = \mathbf{F}_{\text{external}}
    \label{eq:eom}
\end{equation}
where $\mathbf{M}$ is the system's mass matrix, $\mathbf{x}$ is the global vector of node positions, and $\mathbf{F}_{\text{external}}$ encompasses all external loads, such as gravity or fluid forces.

For stiff systems, as is common in structural mechanics, implicit time integration methods are favored for their stability. This requires solving a large, non-linear system of algebraic equations at each time step, typically via a Newton-Raphson method. The primary computational bottleneck in this physics-based approach is the repeated evaluation of the elastic energy's gradient (internal forces) and Hessian (tangent stiffness matrix) required within the non-linear solver. Our work explores a data-driven alternative: learning a direct surrogate for the system's dynamic evolution to bypass these costly iterative calculations entirely.

\begin{figure*}[t]
    \centering
    \includegraphics[width=0.9\textwidth]{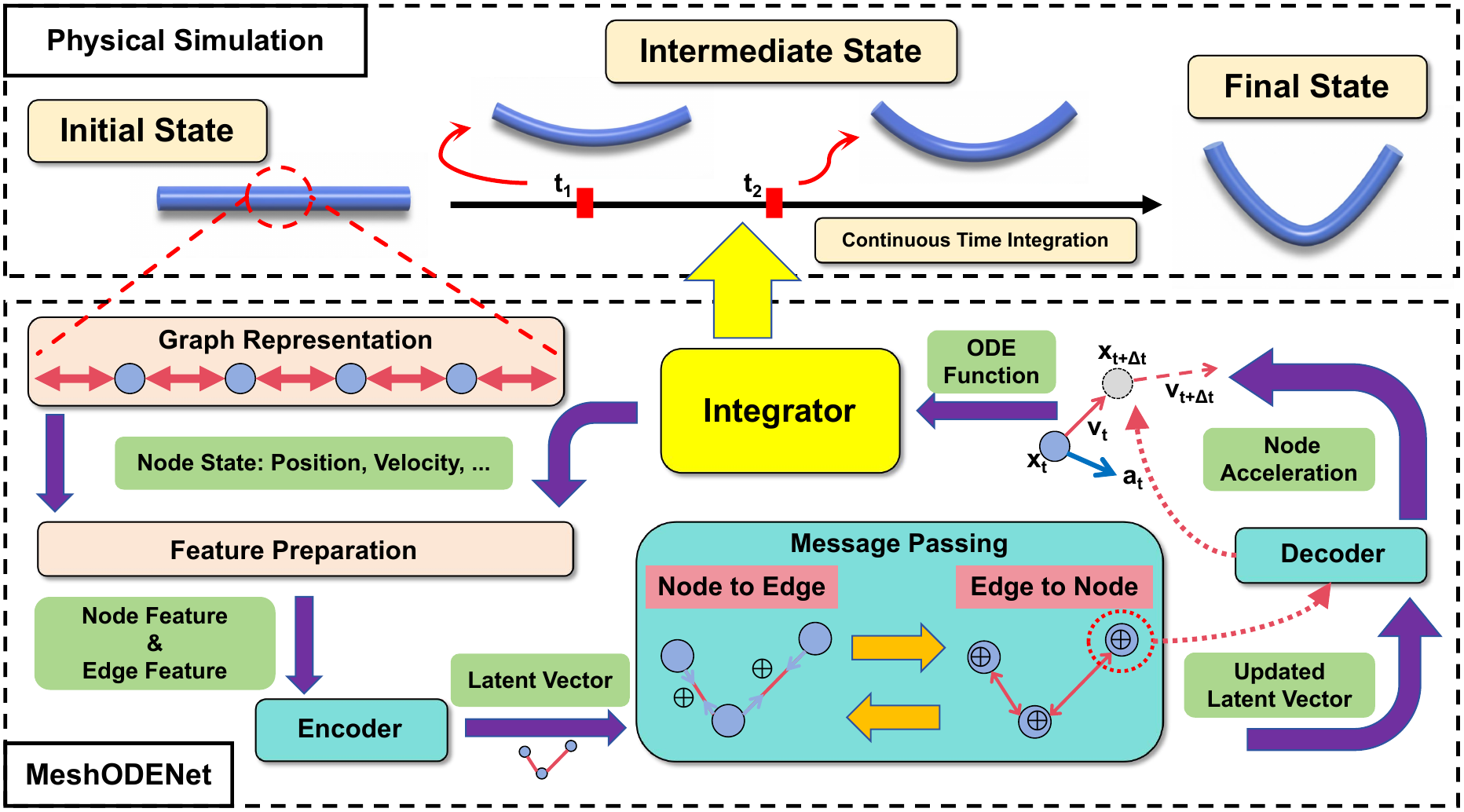}
    \caption{Overall architecture of the MeshODENet framework, illustrating the workflow from graph construction to time integration via the coupled GNN and ODE solver.}
    \label{fig:overall_architecture} 
\end{figure*}

\subsection{The Proposed \textbf{MeshODENet} Framework}
As Fig.~\ref{fig:overall_architecture} shows, \textbf{MeshODENet} learns the continuous-time evolution of a physical system by integrating a GNN-parameterized vector field. The overall operation is defined as:
\begin{equation}
    \mathbf{z}(t_1) = \texttt{ODESolve}(\texttt{GNN}, \mathbf{z}(t_0), t_0, t_1; \mathbf{\theta})
\end{equation}
where $\mathbf{z}(t) = [\mathbf{x}(t), \dot{\mathbf{x}}(t)]^T$ is the augmented state of the system. The framework's components are detailed below. 

\subsubsection{Graph-Based State Representation}
The instantaneous state of the physical system is encoded in a graph $\mathcal{G}_t = (\mathcal{V}, \mathcal{E}, \mathbf{H}_v, \mathbf{H}_e)$, where $\mathcal{V}$ are nodes, $\mathcal{E}$ are edges, and $\mathbf{H}_v, \mathbf{H}_e$ are their respective feature matrices. The feature design is customizable to the specific physics. For the physical systems considered in this work, the interactions are reciprocal; therefore, we model the graph with undirected edges. In the GNN implementation, each undirected edge is represented as a pair of directed edges (one in each direction), ensuring that information flows mutually between connected nodes, consistent with Newton's third law.

\begin{itemize}
    \item \textbf{Node Features:} For node $v_i$, the feature vector $\mathbf{h}_{v_i}$ contains local nodal state. The 1-D beam simulation in \ref{fig:DER} includes the node's velocity, material property (Young's modulus), local geometry, and external forces:
    \begin{equation}
    \label{eq:node_features}
        \mathbf{h}_{v_i} = [\dot{\mathbf{x}}_i, E_i, \cos(\alpha_i), \sin(\alpha_i), \mathbf{F}_{\text{ext},i}]
    \end{equation}
    where $\dot{\mathbf{x}}_i$ is the velocity, $E_i$ is the Young's modulus, $\alpha_i$ is the angle formed by the two edges connected to the node, and $\mathbf{F}_{\text{ext},i}$ is the vector of external forces (gravity and hydrodynamic forces) applied to the node. Here we use both cosine and sine instead of angle value to avoid discontinuities associated with angle wrapping.
    \item \textbf{Edge Features:} For an edge $e_{ij}$ connecting nodes $v_i$ and $v_j$, the feature vector $\mathbf{h}_{e_{ij}}$ encodes both its initial and current geometric states:
    \begin{equation}
    \label{eq:edge_features}
        \mathbf{h}_{e_{ij}} = [l^0_{ij}, \mathbf{d}^0_{ij}, l_{ij}, \mathbf{d}_{ij}]
    \end{equation}
    where $l^0_{ij}$ and $\mathbf{d}^0_{ij}$ are the magnitude and unit direction vector of the edge in its initial, undeformed state, while $l_{ij} = \|\mathbf{x}_j - \mathbf{x}_i\|$ and $\mathbf{d}_{ij} = (\mathbf{x}_j - \mathbf{x}_i) / l_{ij}$ represent the current magnitude and direction. This provides the GNN with explicit information about the edge's deformation.
\end{itemize}

\subsubsection{GNN-Parameterized Dynamics Function}
The core of our framework is a GNN that learns a mapping from the current graph state $\mathcal{G}_t$ to the time derivative of the system's state variables. For second-order dynamical systems, which are described by first-order (e.g., velocity $\dot{\mathbf{x}}$) and zeroth-order (e.g., position $\mathbf{x}$) quantities, our GNN learns to approximate the second-order derivative. In 1-D beam in \ref{fig:DER}, the second-order derivative corresponds to the nodal acceleration, $\mathbf{a}(t) = \ddot{\mathbf{x}}(t)$. The entire GNN operation, which maps the graph state to the acceleration vector, can be expressed as a composition of three functions:
\begin{equation}
    \mathbf{a}(t) = \texttt{GNN}(\mathcal{G}_t; \mathbf{\theta}) = \texttt{Decoder}(\texttt{Processor}(\texttt{Encoder}(\mathcal{G}_t)))
\end{equation}
The learnable parameters of these three components collectively form the GNN parameters $\mathbf{\theta}$.

\paragraph{Encoder} The encoder employs two separate MLPs, $\text{MLP}_{v,enc}$ and $\text{MLP}_{e,enc}$, to project the raw node and edge features into higher-dimensional latent vectors. Specifically, the initial latent representations of node $i$ and edge $(i,j)$ are defined as
\begin{align}
\label{eq:encoder}
    \tilde{\mathbf{h}}_{v_i}^{(0)} = \text{MLP}_{v,enc}(\mathbf{h}_{v_i}), 
    \quad 
    \tilde{\mathbf{h}}_{e_{ij}}^{(0)} = \text{MLP}_{e,enc}(\mathbf{h}_{e_{ij}}),
\end{align}
where the superscript $(0)$ denotes the initial latent state, i.e., the representation before any temporal evolution or iterative message-passing updates.

\paragraph{Processor} The processor refines these latent representations over $L$ message-passing layers. For each layer $l \in \{0, \dots, L-1\}$, it first computes messages $\mathbf{m}_{e_{ij}}$ for each edge based on the states of the edge and its two endpoint nodes. Then, it aggregates the incoming messages for each node and updates the node's representation. This process is governed by learned update functions $\phi_e^{(l)}$ and $\phi_v^{(l)}$, which are implemented as MLPs:
\begin{align}
\label{eq:processor}
    \mathbf{m}_{e_{ij}}^{(l+1)} &= \phi_e^{(l)}(\tilde{\mathbf{h}}_{e_{ij}}^{(l)}, \tilde{\mathbf{h}}_{v_i}^{(l)}, \tilde{\mathbf{h}}_{v_j}^{(l)}) \\
    \tilde{\mathbf{h}}_{v_i}^{(l+1)} &= \phi_v^{(l)}\left(\bigoplus_{v_j \in \mathcal{N}(v_i)} \mathbf{m}_{e_{ji}}^{(l+1)}, \tilde{\mathbf{h}}_{v_i}^{(l)}\right)
\end{align}
where $\mathcal{N}(v_i)$ is the set of neighbors of node $v_i$, and $\bigoplus$ is a permutation-invariant aggregation operator, such as summation.

\paragraph{Decoder} After the final message-passing layer, a decoder MLP, $\text{MLP}_{dec}$, maps the final latent representation of each node, $\tilde{\mathbf{h}}_{v_i}^{(L)}$, back to the target physical quantity. For our structural examples, this is the predicted acceleration $\mathbf{a}_i$:
\begin{equation}
\label{eq:decoder}
    \mathbf{a}_i = \text{MLP}_{dec}(\tilde{\mathbf{h}}_{v_i}^{(L)})
\end{equation}

\subsubsection{Coupling Integrator}
The GNN architecture described in the previous section provides the core function for parameterizing the system's dynamics. The collective output of GNN is used to define the vector field for a system of first-order ODEs. For second-order mechanical systems, this leads to the following formulation for the augmented state $\mathbf{z}(t) = [\mathbf{x}(t), \dot{\mathbf{x}}(t)]^T$, with the initial condition $\mathbf{z}(0) = [\mathbf{x}(0), \dot{\mathbf{x}}(0)]^T$:
\begin{equation}
    \frac{d\mathbf{z}(t)}{dt} = \frac{d}{dt}
    \begin{bmatrix}
        \mathbf{x}(t) \\
        \dot{\mathbf{x}}(t)
    \end{bmatrix}
    =
    \begin{bmatrix}
        \dot{\mathbf{x}}(t) \\
        \mathbf{a}(t)
    \end{bmatrix}
    =
    \begin{bmatrix}
        \dot{\mathbf{x}}(t) \\
        \texttt{GNN}(\mathcal{G}_t; \mathbf{\theta})
    \end{bmatrix}
    \label{eq:node_system_final}
\end{equation}

The solution to this ODE, which gives the state at a future time $t_1$, is formally given by the integral:

\begin{equation}
    \mathbf{z}(t_1) = \mathbf{z}(t_0) + \int_{t_0}^{t_1} \frac{d\mathbf{z}(\tau)}{dt} \,d\tau = 
    \begin{bmatrix}
        \mathbf{x}(t_0)\\
        \dot{\mathbf{x}}(t_0)
    \end{bmatrix}
    + \int_{t_0}^{t_1} 
    \begin{bmatrix}
        \dot{\mathbf{x}}(\tau) \\
        \texttt{GNN}(\mathcal{G}_\tau; \mathbf{\theta})
    \end{bmatrix}
    \,d\tau
    \label{eq:integral_form}
\end{equation}

To solve this initial value problem and predict the system's state over time, a numerical ODE solver is required. A variety of numerical schemes are available for this purpose, ranging from simple forward Euler methods to more sophisticated adaptive step-size solvers like the Dormand-Prince pair. In this work, we employ the widely-used explicit fourth-order Runge-Kutta (RK4) method. This fixed-step solver offers a robust balance between computational efficiency and accuracy for many dynamical systems, and it requires evaluating the GNN function four times per integration step. 

This explicit coupling of the GNN and the integrator ensures that the dynamics are continuously informed by the current physical state of the graph $\mathcal{G}_t$. This formulation is also flexible; for other physics, the GNN could be trained to learn the other field's rate of change, leading to a different but structurally similar ODE system.

\subsubsection{Loss Function}
The GNN parameters $\mathbf{\theta}$ are learned by minimizing a loss function $\mathcal{L}$ that is calculated on the predicted trajectory. Although our GNN predicts a second-order quantity (acceleration), we define the loss on the integrated, zeroth-order quantity (position). Supervising the model on the integrated trajectory provides a more stable training signal and better enforces long-term physical consistency, as it implicitly penalizes the accumulation of errors in both velocity and acceleration over time. This principle is generalizable: for other physical systems, defining the loss on the primary state for the variable of interest (e.g., temperature, concentration) is often more effective than supervising its derivatives directly. The loss function is thus defined as the mean squared error between the predicted and ground-truth position trajectories over $N_p$ time steps for a batch of $N_s$ samples:
\begin{equation}
    \mathcal{L}(\mathbf{\theta}) = \frac{1}{N_s N_p} \sum_{i=1}^{N_s} \sum_{j=1}^{N_p} \| \hat{\mathbf{x}}^{(i)}(t_j) - \mathbf{x}^{(i)}(t_j) \|^2_2
    \label{eq:loss_function_final}
\end{equation}
where $\hat{\mathbf{x}}^{(i)}(t_j)$ is the predicted position trajectory obtained by integrating Eq.~\eqref{eq:node_system_final}. The gradient of the loss with respect to $\mathbf{\theta}$ is computed efficiently by backpropagating through the ODE solver via the adjoint sensitivity method.

\subsubsection{Training Details}

We implement our framework using PyTorch, with graph operations handled by the PyTorch Geometric (PyG) library. 
During both training and inference, the model takes an initial state from a ground-truth trajectory and evolves it forward in continuous time using a Neural ODE solver, which integrates the learned dynamics function over the entire time horizon. 
The loss is then computed on the full rollout against the reference trajectory. 
This approach differs from autoregressive prediction, where the model would generate the trajectory step by step, repeatedly feeding its own one-step prediction back as input. 
By contrast, our Neural ODE formulation treats the system as a continuous dynamical process and directly integrates its evolution, avoiding the compounding of step-wise errors typical of autoregressive schemes.

\section{Experiment and Results Discussion}
\label{sec:experiments}

In this section, we present a series of numerical experiments designed to rigorously evaluate the performance of our proposed \textbf{MeshODENet} framework. We validate its accuracy, long-term stability, and generalization capabilities across a range of challenging structural mechanics problems. Specifically, we assess the model on one- and two-dimensional elastic bodies undergoing large, non-linear deformations, which serve as canonical tests for computational mechanics solvers.

\begin{figure}
    \centering
    \includegraphics[width=1\linewidth]{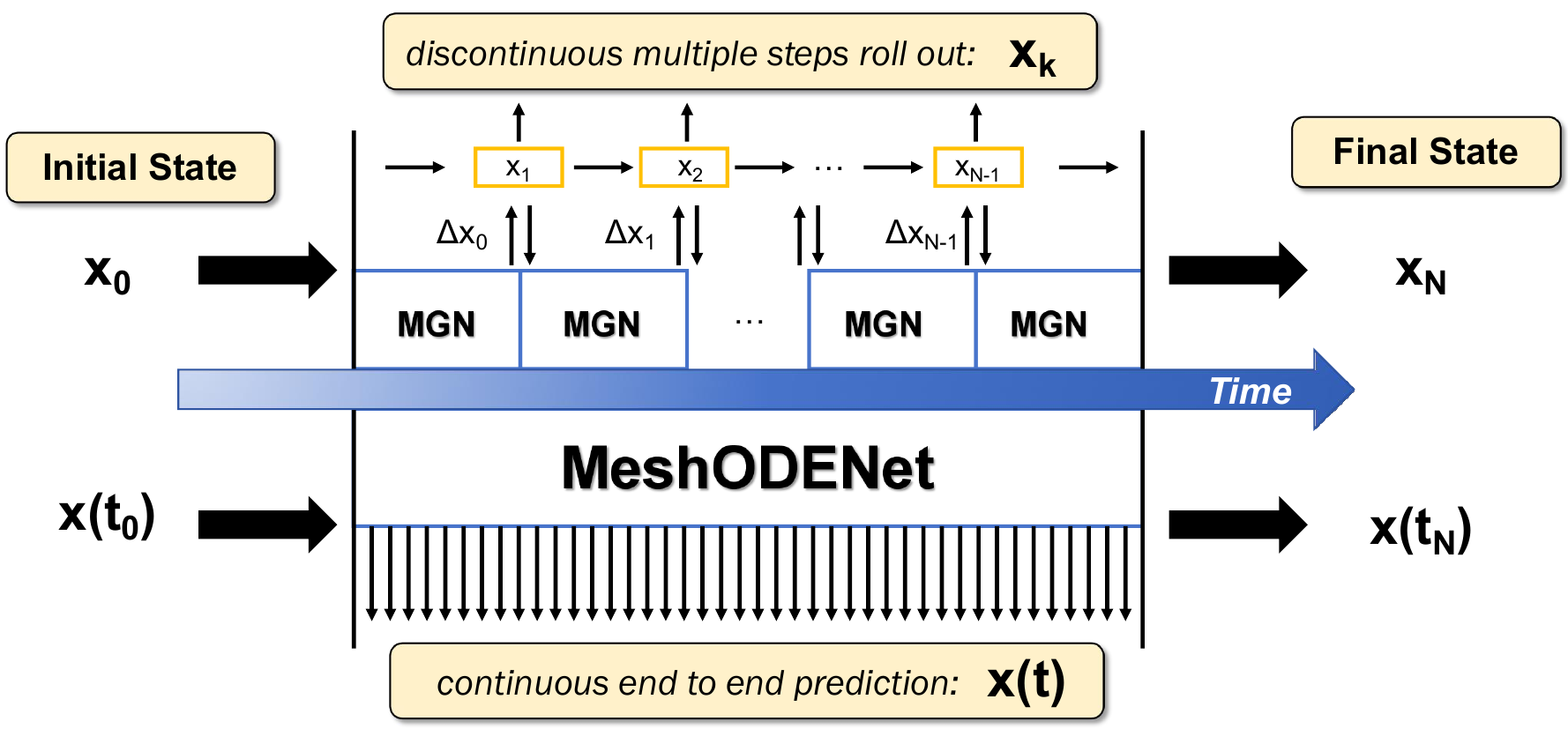}
    \caption{Conceptual illustration of the two modeling frameworks over time. The top path represents the standard autoregressive MGN model, which iteratively predicts the system's state in discrete steps ($\Delta t$). The bottom path represents our proposed \textbf{MeshODENet}, which uses the GNN as a dynamics function within an ODE solver to compute the system's state over a continuous time interval.}
    \label{fig:comparison}
\end{figure}

To provide a comprehensive comparison, we evaluate our model against MeshGraphNet (MGN), a standard GNN framework that operates in an autoregressive, single-step manner\cite{Pfaff2021}. The fundamental difference in their temporal modeling is conceptually illustrated in Fig.~\ref{fig:comparison}. As depicted in the figure, the MGN model performs a series of discrete, sequential predictions, where each step's output becomes the input for the next. In contrast, our \textbf{MeshODENet} formulates the problem continuously, using an ODE solver to integrate the system's state from an initial time to a future point. This direct comparison in the subsequent experiments allows us to isolate and quantify the benefits of this continuous-time formulation.

\subsection{Case Study 1: 1-D Elastic Rod Falling in Fluid}

\subsubsection{Problem Setup}

The first experiment validates the fundamental capabilities of our framework on the motivating example introduced in Sec.~\ref{sec:methodology}: a 1-D elastic beam falling under gravity in a viscous fluid. 
The ground-truth trajectories are generated using a high-fidelity Discrete Elastic Rods (DER) solver, where the beam of length $L=0.10$~m is discretized into $n_v=21$ vertices with segment length $\Delta L = L/(n_v-1)$. 
Each vertex is assigned a lumped mass corresponding to a sphere of radius $R=\Delta L/10$ (with the central node modeled by a larger effective radius $R_{\mathrm{mid}}$), yielding a total mass distribution consistent with the rod geometry and density. Internal elastic forces are derived from axial stretching/compression and bending energies, parameterized respectively by

\begin{equation}
EA = Y \pi r_0^2, \qquad EI = \tfrac{1}{4} Y \pi r_0^4,
\end{equation}

where $r_0 = 1$~mm is the rod radius and $Y$ is the Young's modulus. 
External loading consists of (i) gravity minus buoyancy, implemented as $\mathbf{F}_{g,i} = (\rho_{\text{metal}}-\rho_{\text{fluid}}) V_i \mathbf{g}$ for each node, and (ii) low-Reynolds-number viscous drag, modeled as a Stokes-type damping force

\begin{equation}
\mathbf{F}_{\mathrm{drag},i} = - 6 \pi \mu a_i \mathbf{v}_i,
\end{equation}

with $\mu=1000$~Pa$\cdot$s denoting fluid viscosity, $a_i$ the effective nodal radius, and $\mathbf{v}_i$ the nodal velocity. 
Because gravity acts vertically, the deformation is predominantly in-plane, and in the discrete formulation this bending response is captured by the stretching/compression of nonlinear springs between adjacent nodes.

We created a dataset of 71 distinct trajectories for training, where each simulation features a beam with a different sampled Young's modulus $E$ drawn from a uniform distribution over $[0.1,\,1.5]$~GPa. 
For testing, 14 additional trajectories are generated with $E$ sampled from an overlapping range $[0.13,\,1.43]$~GPa to evaluate the model's generalization capabilities. 
Each trajectory spans 20~s with a sampling interval of 0.1~s. 
The models are tasked with predicting the entire trajectory from the initial state.

\begin{table*}[htp!]
    \centering
    \caption{Mean and standard deviation of RMSE for the 1-D elastic rod simulation, averaged over 14 test trajectories. The values in brackets denote the relative percentage increase in Mean RMSE of the MGN model compared to MeshODENet.}
    \label{tab:prediction_error_1D}
    \begin{tabular}{l rrrrrrr}
        \hline\hline
        Prediction & \multicolumn{7}{c}{RMSE  ($\times 10^{-4}$)} \\
        \cline{2-8} 
        \addlinespace 
        Step & \multicolumn{1}{c}{1} & \multicolumn{1}{c}{30} & \multicolumn{1}{c}{60} & \multicolumn{1}{c}{90} & \multicolumn{1}{c}{120} & \multicolumn{1}{c}{150} & \multicolumn{1}{c}{180}\\
        \hline
        \addlinespace 
        MeshODENet & 2.62$\pm$0.13 & 5.79$\pm$5.57 & 4.61$\pm$6.95 & 3.05$\pm$4.66 & 1.89$\pm$2.32 & 1.62$\pm$1.61 & 2.03$\pm$1.86 \\
        \addlinespace 
        MGN & 4.14$\pm$0.02 & 19.74$\pm$15.55 & 28.32$\pm$23.96 & 37.98$\pm$31.63 & 64.71$\pm$54.82 & 116.53$\pm$94.85 & 192.58$\pm$146.94 \\
        & (57.85\%) & (240.98\%) & (514.20\%) & (1146.61\%) & (3325.11\%) & (7092.38\%) & (9370.11\%)\\
        \addlinespace 
        \hline\hline
    \end{tabular}
    \vspace{0.5em} 
    \small 
\end{table*}

\subsubsection{Parameters Setting}

We compare our \textbf{MeshODENet} against a standard autoregressive MGN model. Both models share a hidden dimension of 128. Based on our preliminary experiments, we set the number of message-passing layers for the MGN model to 15 to achieve its optimal performance. While frameworks incorporating Neural ODEs are often associated with high memory usage and long training times, our preliminary experiments revealed that \textbf{MeshODENet} can achieve excellent predictive accuracy with just a single message-passing layer. This is likely because the iterative nature of the ODE solver handles the temporal information propagation, allowing the GNN to focus solely on spatial message passing. This significant architectural simplification makes our model highly efficient in terms of both training speed and memory consumption.

\begin{figure}[htp!]
    \centering
    \includegraphics[width=1\linewidth]{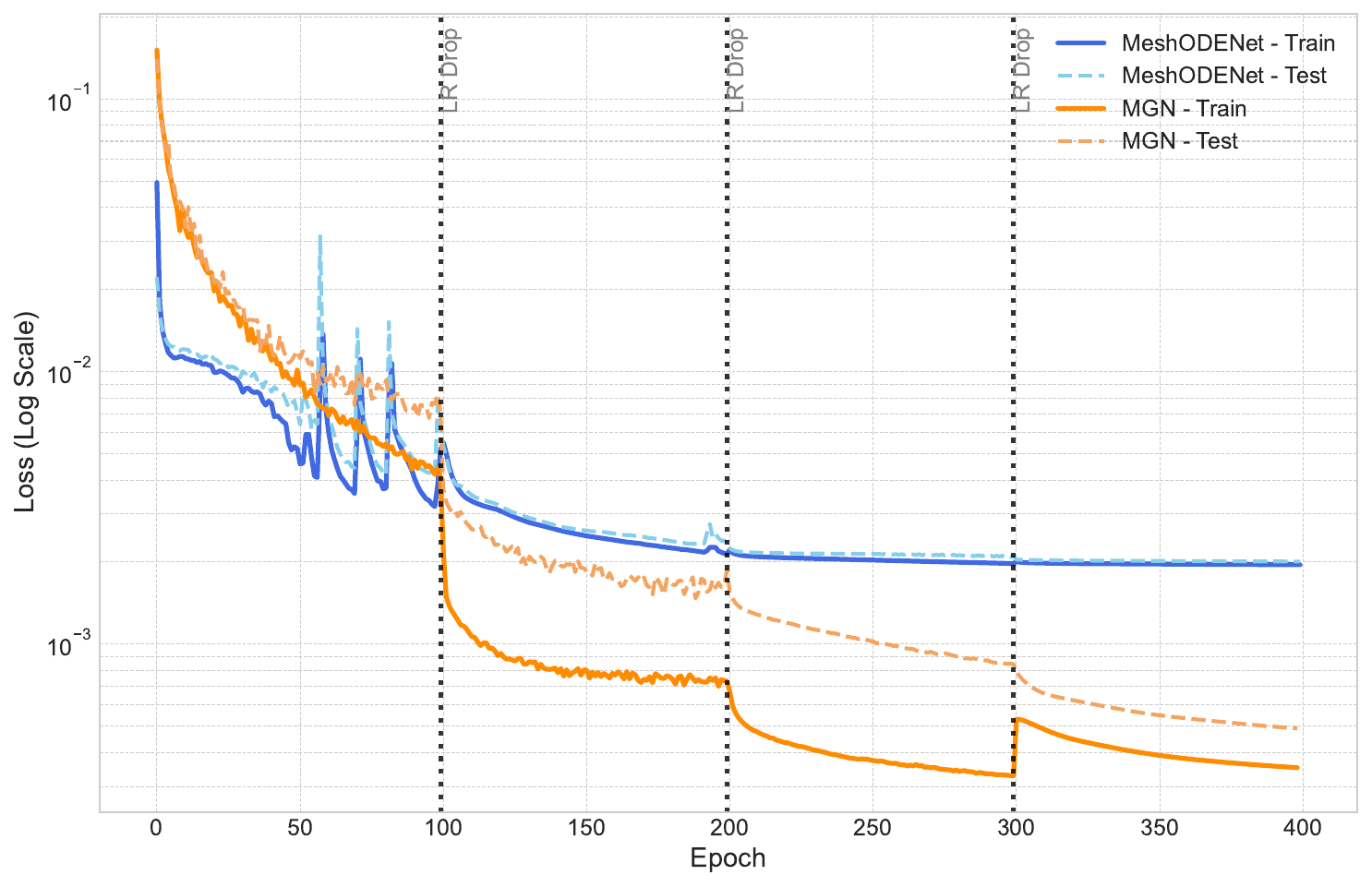}
    \caption{Training and test loss curves of MGN and \textbf{MeshODENet} over 400 epochs.}
    \label{fig:loss_curves_1}
\end{figure}

Both models were trained for 400 epochs using the Adam optimizer. The initial learning rate was set to $1 \times 10^{-4}$ and decayed by a factor of 10 every 100 epochs, with a weight decay of $5 \times 10^{-4}$ applied for regularization. The corresponding training and test loss curves are shown in Fig.~\ref{fig:loss_curves_1}, where the scheduled learning rate drops are also indicated. It should be noted that a direct comparison of the absolute loss values is not meaningful, since MGN is supervised on single-step predictions while \textbf{MeshODENet} is trained on entire trajectory rollouts. Therefore, the relative performance of the two frameworks is evaluated in terms of predictive accuracy on the test set, as detailed in the following section.

\begin{figure*}[htp!]
    \centering

    \begin{subfigure}[b]{0.95\textwidth}
        \centering
        \includegraphics[width=\textwidth]{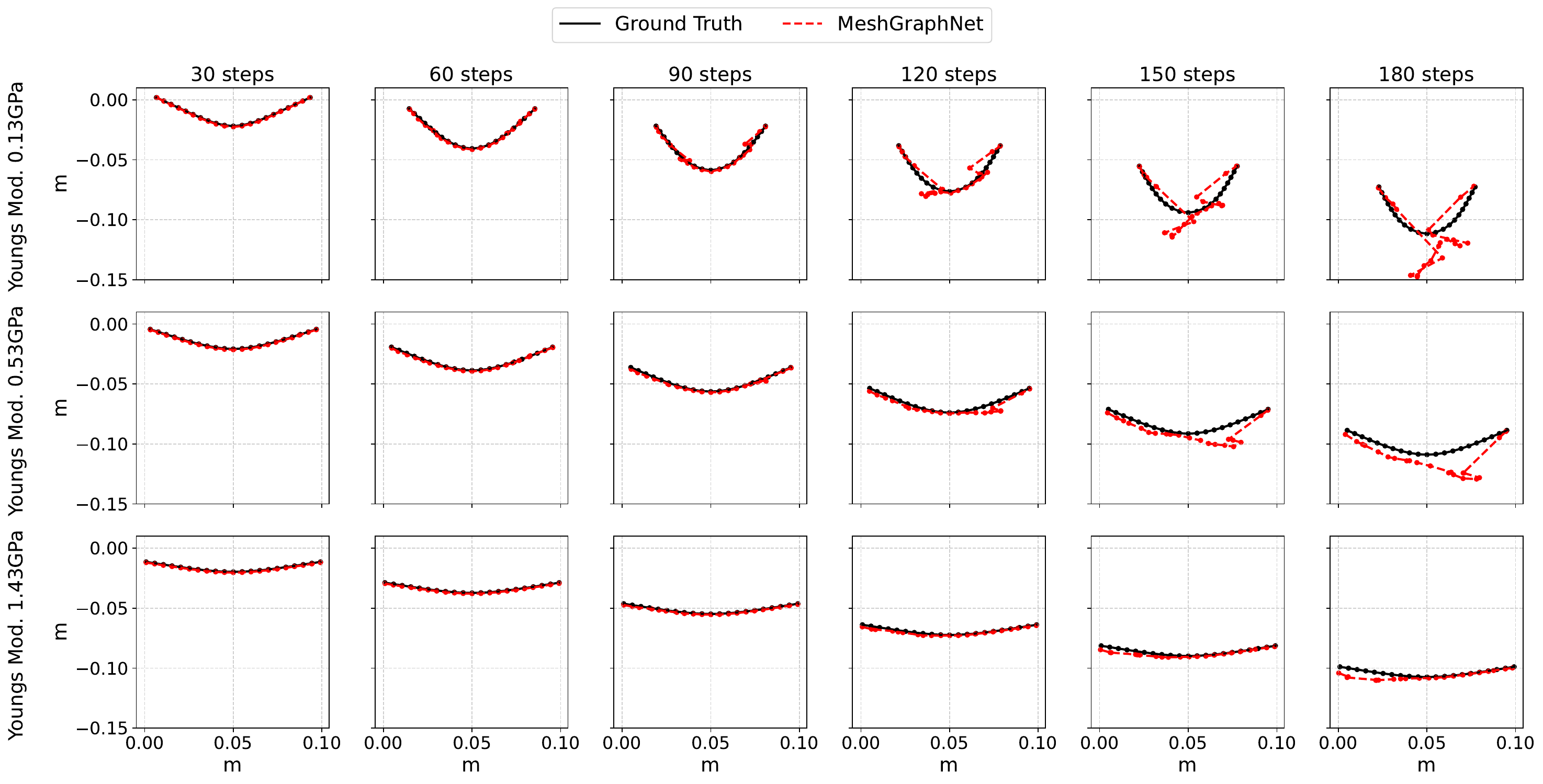}
        \caption{MGN Predictions}
        \label{fig:1D_noode}
    \end{subfigure}
    
    \vspace{1em}
    
    \begin{subfigure}[b]{0.95\textwidth}
        \centering
        \includegraphics[width=\textwidth]{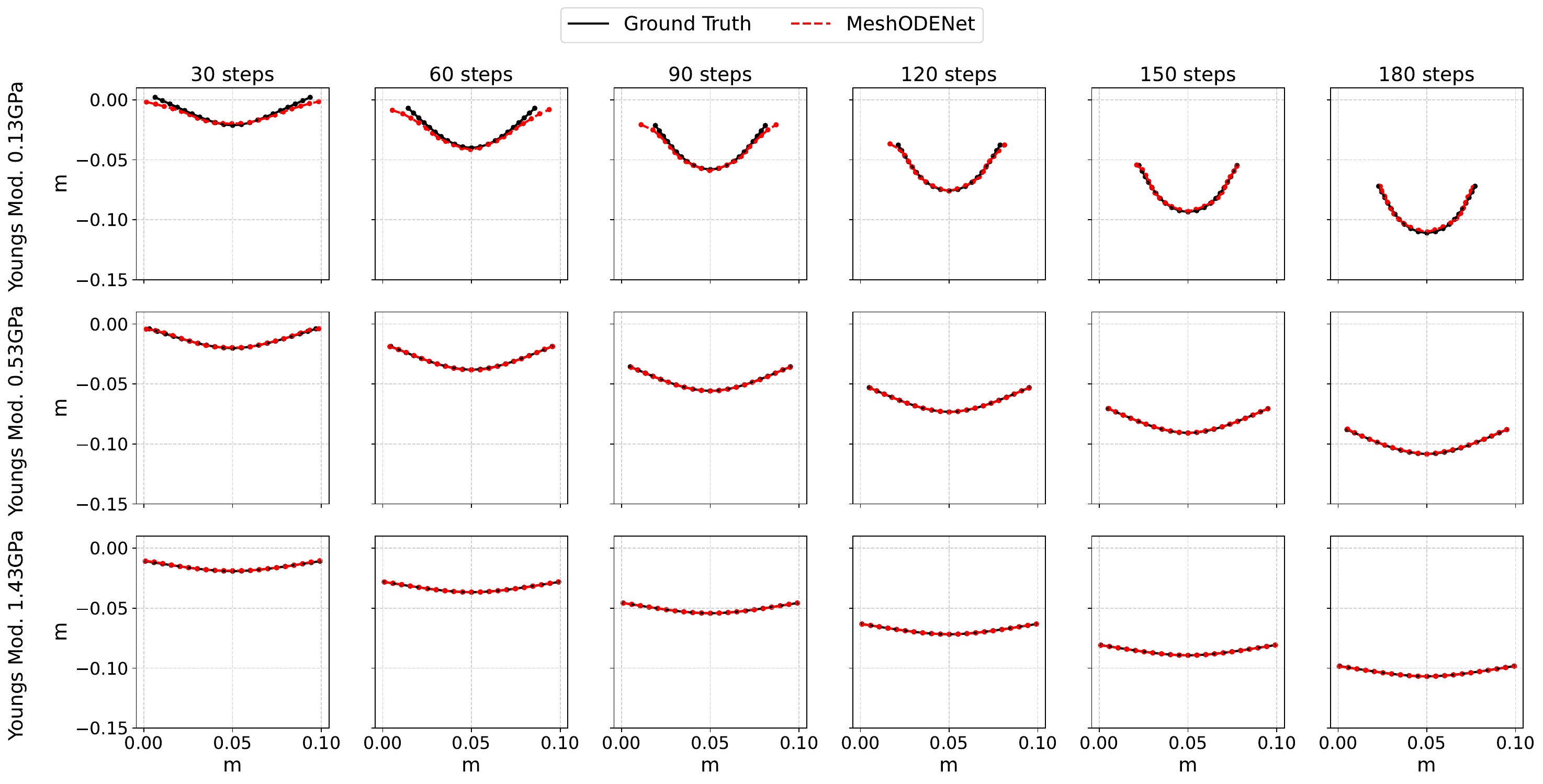}
        \caption{MeshODENet Predictions (Our Model)}
        \label{fig:1D_ode}
    \end{subfigure}
    
    \caption{Qualitative comparison of rollout predictions for the 1-D elastic rod on three test cases with different Young's moduli. Each column represents a specific time step. (a) The autoregressive MGN model shows error accumulation leading to instability. (b) Our MeshODENet maintains long-term stability and physical plausibility.}
    \label{fig:1D_comparison}
\end{figure*}

\subsubsection{Results and Discussion}

The quantitative performance of both models on the 14 test trajectories is summarized in Table~\ref{tab:prediction_error_1D}. The table presents the Root Mean Square Error (RMSE), averaged over all test cases, at various prediction time steps. The values in brackets indicate the relative percentage increase in error of the MGN model compared to our \textbf{MeshODENet}, highlighting the performance gap between the two approaches.

Qualitative results for three representative test cases with different Young's moduli are presented in Fig.~\ref{fig:1D_comparison}. Fig.~\ref{fig:1D_comparison}(a) shows the predictions from the MGN model, while Fig.~\ref{fig:1D_comparison}(b) displays the results from our \textbf{MeshODENet}.

From these results, a clear performance trend emerges. The MGN model (Fig.~\ref{fig:1D_comparison}a) makes reasonably accurate predictions during the initial, shorter time steps. However, as the prediction horizon extends, the autoregressive nature of the model leads to a rapid accumulation of errors. This is particularly evident in the more flexible cases (e.g., Young's Mod. 0.13 GPa), where the beam's structure is completely destroyed at later time steps, resulting in physically implausible configurations. This instability is quantitatively reflected in Table~\ref{tab:prediction_error_1D}, where the RMSE for MGN grows exponentially, increasing by over 9000\% relative to our model by the 180th step.

In stark contrast, our \textbf{MeshODENet} (Fig.~\ref{fig:1D_comparison}b) demonstrates exceptional long-term stability across all test cases. The predicted trajectories remain physically coherent and closely match the ground truth throughout the entire rollout. It is important to note that while the figure displays snapshots at discrete time steps for comparison, the underlying prediction from our model is continuous, generated by the ODE solver.

However, a nuanced analysis also reveals a limitation and a direction for future improvement. Because \textbf{MeshODENet} predicts the entire trajectory based solely on the initial state, it can sometimes exhibit a delayed response to sudden, violent changes in dynamics. This is most noticeable in the initial, highly dynamic phase of the most flexible beam (Young's Mod. 0.13 GPa), where MGN's single-step prediction can appear momentarily more accurate. This suggests that while our framework excels in long-term stability, future work could explore hybrid approaches to improve its responsiveness to sharp, transient events.

\begin{table*}[htp!]
    \centering
    \caption{Mean and standard deviation of RMSE for the 2-D cantilever plate simulation, averaged over 6 test trajectories. The values in brackets denote the relative percentage increase in Mean RMSE of the MGN model compared to MeshODENet.}
    \label{tab:prediction_error_2D}
    \begin{tabular}{l rrrrrr}
        \hline\hline
        Prediction & \multicolumn{6}{c}{RMSE  ($\times 10^{-4}$)} \\
        \cline{2-7} 
        \addlinespace 
        Step & \multicolumn{1}{c}{1} & \multicolumn{1}{c}{25} & \multicolumn{1}{c}{50} & \multicolumn{1}{c}{150} & \multicolumn{1}{c}{250} & \multicolumn{1}{c}{350}\\
        \hline
        \addlinespace 
        MeshODENet & 0.32$\pm$0.00 & 8.29$\pm$0.31 & 6.84$\pm$0.58 & 3.11$\pm$0.72 & 2.49$\pm$0.74 & 2.72$\pm$0.32  \\
        \addlinespace 
        MGN & 1.73$\pm$0.02 & 19.61$\pm$6.46 & 60.87$\pm$26.89 & 188.06$\pm$29.04 & 267.20$\pm$41.90 & 354.18$\pm$100.43  \\
        & (440.62\%) & (136.55\%) & (789.91\%) & (5946.95\%) & (10630.92\%) & (12873.63\%) \\
        \addlinespace 
        \hline\hline
    \end{tabular}
    \vspace{0.5em} 
    \small 
\end{table*}

\begin{figure*}[htp!]
    \centering

    \begin{subfigure}[b]{0.95\textwidth}
        \centering
        \includegraphics[width=\textwidth]{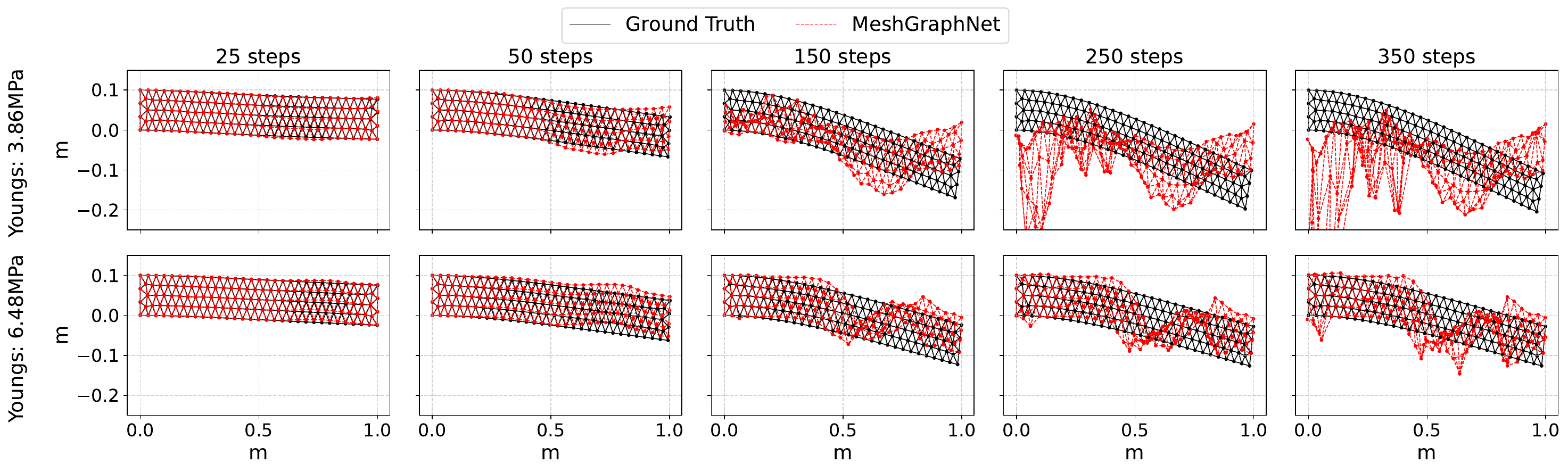}
        \caption{MGN Predictions}
        \label{fig:2D_noode}
    \end{subfigure}
    
    \vspace{1em}
    
    \begin{subfigure}[b]{0.95\textwidth}
        \centering
        \includegraphics[width=\textwidth]{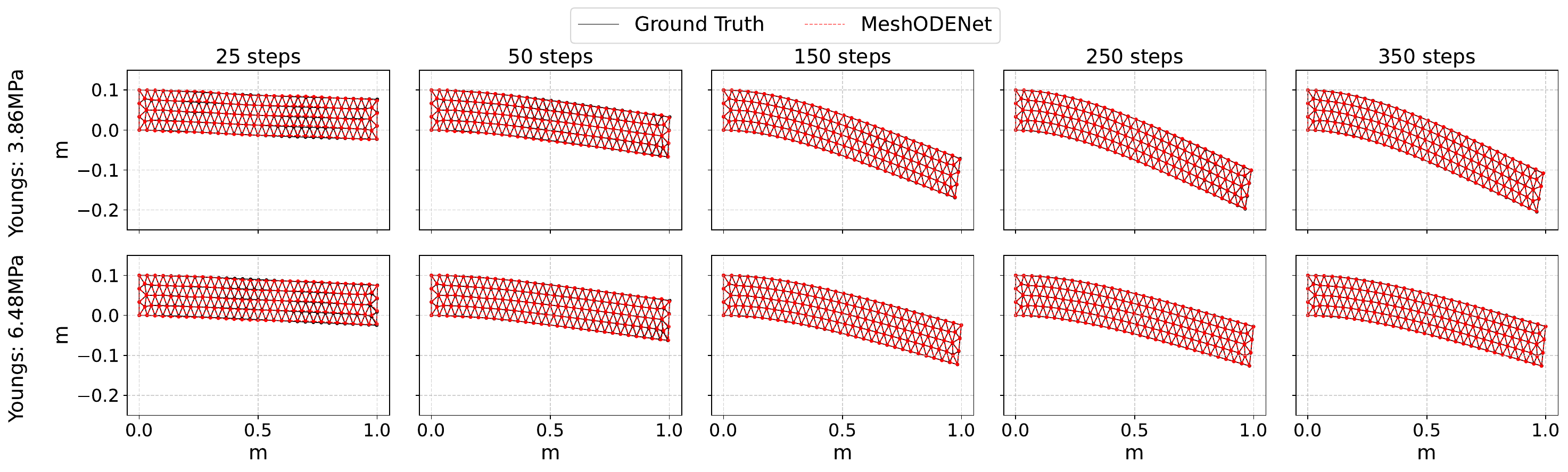}
        \caption{MeshODENet Predictions (Our Model)}
        \label{fig:2D_ode}
    \end{subfigure}
    
    \caption{Qualitative comparison of rollout predictions for the 2-D cantilever plate on two test cases with different Young's moduli. (a) The autoregressive MGN model rapidly becomes unstable and loses physical plausibility. (b) Our MeshODENet accurately captures the complex, large-deformation behavior over the long term.}
    \label{fig:2D_comparison}
\end{figure*}

\subsection{Case Study 2: Large Deformation of a 2-D Cantilever Plate}

\subsubsection{Problem Setup}

To further test the framework's ability to handle severe geometric nonlinearity, we simulate a 2-D elastic cantilever plate fixed at one end and deforming under gravity. The ground-truth data are generated using a Discrete Elastic Plates (DEP) method \citep{DEP1,DEP2}, which extends the DER formulation to thin plates. The plate has length $L=1$~m, width $w=0.1$~m, thickness $h=0.001$~m, and density $\rho=1200$~kg/m$^3$. External loading consists of gravity acting vertically downward together with a mild velocity-proportional damping term, while the internal elastic response is captured through stretching and bending of the triangular mesh elements. Under vertical loading, the plate undergoes large out-of-plane deformation, which is faithfully represented by the discrete plate formulation.

The dataset consists of 30 training trajectories with Young’s moduli uniformly sampled from $[2.5,\,7.5]$~MPa, and 6 test trajectories sampled from the same interval. Each trajectory is simulated for $1.5$~s, comprising 500 time steps with a step length of $0.003$~s. Compared to the 1-D beam case, the node features for this problem are adapted by removing the angle feature and introducing a \emph{node-type} feature to distinguish fixed boundary nodes from free nodes.

\subsubsection{Parameters Setting}

\begin{figure}[htp!]
    \centering
    \includegraphics[width=1\linewidth]{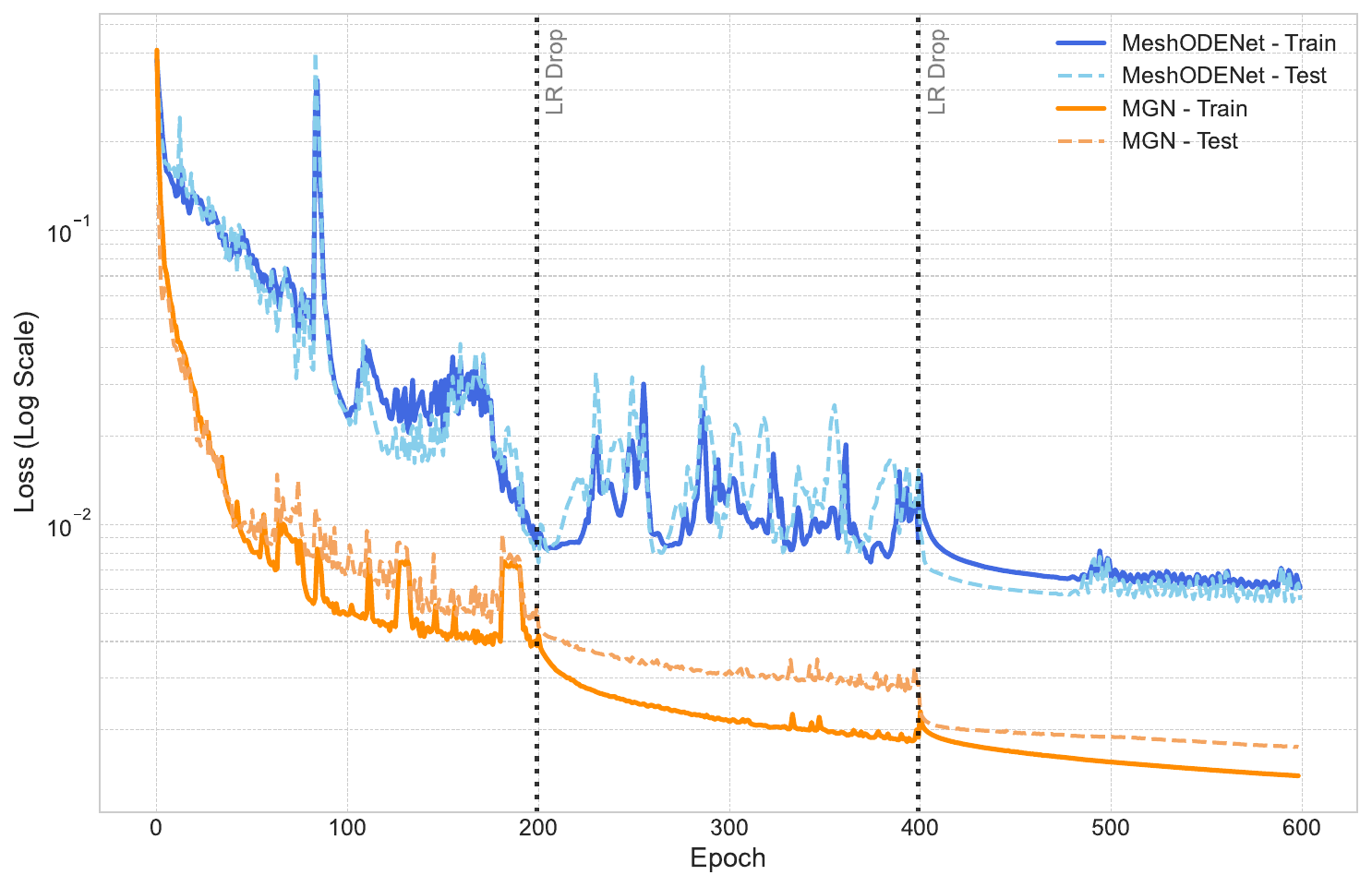}
    \caption{Training and test loss curves of MGN and \textbf{MeshODENet} over 600 epochs.}
    \label{fig:loss_curves_2}
\end{figure}

The hyperparameter configuration was kept consistent with the previous case study. For the MGN model, a depth of 30 message-passing layers was used, as identified from preliminary experiments. Our \textbf{MeshODENet} continues to employ a single message-passing layer, with temporal evolution handled by the ODE solver. Both models were trained for 600 epochs with the Adam optimizer, starting from a learning rate of $1 \times 10^{-4}$, which was decayed by a factor of 10 at epochs 200 and 400. The corresponding training and test loss curves are shown in Fig.~\ref{fig:loss_curves_2}.

\subsubsection{Results and Discussion}

The quantitative results, averaged over the six test trajectories, are presented in Table~\ref{tab:prediction_error_2D}. The table details the Mean RMSE at various time steps and the corresponding percentage increase in error for the MGN model relative to \textbf{MeshODENet}. The qualitative rollout predictions for two representative test cases are visualized in Fig.~\ref{fig:2D_comparison}.

The results from this more complex 2-D scenario reinforce the conclusions from the first case study. As shown in Table~\ref{tab:prediction_error_2D}, the MGN model's error accumulates at an even more dramatic rate, a consequence of the increased complexity and larger deformations of the 2-D plate. The qualitative results in Fig.~\ref{fig:2D_comparison}(a) vividly illustrate this failure mode: while MGN captures the initial downward motion, it quickly loses structural integrity, leading to chaotic and physically incorrect predictions.

In contrast, our \textbf{MeshODENet} once again demonstrates better long-term stability (Fig.~\ref{fig:2D_comparison}(b)). More importantly, it accurately captures the whole phases of the plate's motion. This ability to distinguish between different dynamic regimes highlights the strength of the continuous-time formulation in learning a robust representation of the underlying physical laws, making it a more reliable simulator for complex, non-linear systems.

\section{Limitations and Future Work}
\label{sec:Limitations}

While our proposed framework demonstrates promising results in terms of accuracy, stability, and generalization, we acknowledge several limitations that pave the way for future research. The primary limitation of the current framework lies in the substantial memory consumption associated with graph data, particularly for systems with a large number of nodes. This can increase the computational cost of model training, posing a challenge for extremely large-scale problems.

Therefore, a key direction for future work is to enhance the computational efficiency and scalability of the model. One promising avenue is to explore methods for learning from localized sub-domain samples, enabling the model to predict the behavior of an entire complex system without requiring the full graph in memory. This approach could significantly reduce the computational cost while maintaining predictive accuracy.

Furthermore, we aim to extend the framework's capabilities to handle more complex physical interactions. A key challenge in this context is the high degrees of freedom inherent in many physical systems, which can lead to sparse training data when complex external forces are considered. This necessitates the meticulous design of input features and the incorporation of stronger physical priors to ensure that the model can effectively learn the underlying force-response mechanisms. Addressing these challenges will be crucial for applying our framework to real-time control of complex dynamical systems, opening up broader applications in robotics, biomechanics, and virtual engineering.

\section{Conclusion}
\label{sec:Conclusion}

In this paper, we introduced \textbf{MeshODENet}, a general and versatile framework that synergistically combines Graph Neural Networks with Neural Ordinary Differential Equations for the long-term, stable simulation of complex physical systems. We identified a critical gap in the existing literature, where GNODE-based approaches have been underutilized in computational mechanics, and existing architectures often lack the physical fidelity required for evolving, tightly-coupled systems.

Our proposed framework addresses these shortcomings by embedding a GNN, which operates on the current physical state of the system, directly as the vector field within a continuous-time ODE solver. Through a series of challenging case studies on one- and two-dimensional elastic structures undergoing large deformations, we demonstrated that our framework significantly outperforms relevant baselines in long-term prediction accuracy and stability.

The results confirm that \textbf{MeshODENet} provides a powerful new approach for developing data-driven surrogate models in computational mechanics. By offering a general-purpose solution that remains faithful to the underlying physics, this work paves the way for accelerating the design and analysis of complex engineering systems and opens new possibilities for data-driven control and optimization.

\section*{Code availability}
The GitHub code is available at \url{https://github.com/leixinma/MeshODENet}.

\section*{Acknowledgements}
L.M. acknowledges the support for her startup funding from Ira A. Fulton Schools of Engineering at Arizona State University, support from USDA Award No.2024-67021-42526, and Early Career Fellowship in Gulf Research Program from the National Academy of Sciences, Engineering, and Medicine.




\bibliographystyle{asmejour}   

\bibliography{asmejour-sample} 



\end{document}